\documentclass[review]{elsarticle}

\usepackage{hyperref}
\usepackage{lineno}
\usepackage{algorithm}
\usepackage{algpseudocode}
\usepackage{amsmath}
\usepackage{amsfonts}
\usepackage{siunitx}
\usepackage{soul}
\usepackage{xcolor}
\newcommand{\mycomment}[1]{}
\usepackage{booktabs}

\begin{document}

\begin{frontmatter}

\title{Grower-in-the-Loop Interactive Reinforcement Learning for Greenhouse Climate Control}

\author{Maxiu Xiao\fnref{myfootnote1}}
\author{Jianglin Lan\fnref{myfootnote2}}
\author{Jingxin Yu\fnref{myfootnote1,myfootnote3,myfootnote4}}
\author{Weihong Ma\fnref{myfootnote5}}
\author{Qiuju Xie\fnref{myfootnote6}}
\author{Congcong Sun\fnref{myfootnote1}\texorpdfstring{\corref{cor1}}{}}

\fntext[myfootnote1]{Agricultural Biosystems Engineering Group, Wageningen University, 6700 AA Wageningen, The Netherlands}
\fntext[myfootnote2]{James Watt School of Engineering, University of Glasgow, Glasgow G12
8QQ, United Kingdom}
\fntext[myfootnote3]{National Engineering Research Center for Intelligent Equipment in Agriculture, Beijing, 100097, China}
\fntext[myfootnote4]{Research Center for Intelligent Equipment Technology, Beijing Academy of Agriculture and Forestry Sciences, Beijing, 100097, China}
\fntext[myfootnote5]{Information Technology Research Center, Beijing Academy of Agriculture and Forestry Sciences, Beijing, 100097, China}
\fntext[myfootnote6]{College of Electrical and Information, Northeast Agricultural University,150030 Harbin, China}
\cortext[cor1]{Corresponding email: congcong.sun@wur.nl}

\begin{abstract}
Climate control is crucial for greenhouse production as it directly affects crop growth and resource use. Reinforcement learning (RL) has received increasing attention in this field, but still faces challenges, including limited training efficiency and high reliance on initial learning conditions. Interactive RL, which combines human (grower) input with the RL agent’s learning, offers a potential solution to overcome these challenges. However, interactive RL has not yet been applied to greenhouse climate control and may face challenges related to imperfect inputs. Therefore, this paper aims to explore the possibility and performance of applying interactive RL with imperfect inputs into greenhouse climate control, by: (1) developing three representative interactive RL algorithms tailored for greenhouse climate control (reward shaping, policy shaping and control sharing); (2) analyzing how input characteristics are often contradicting, and how the trade-offs between them make grower's inputs difficult to perfect; (3) proposing a neural network-based approach to enhance the robustness of interactive RL agents under limited input availability; (4) conducting a comprehensive evaluation of the three interactive RL algorithms with imperfect inputs in a simulated greenhouse environment. The demonstration shows that interactive RL incorporating imperfect grower inputs has the potential to improve the performance of the RL agent. RL algorithms that influence action selection, such as policy shaping and control sharing, perform better when dealing with imperfect inputs, achieving 8.4\% and 6.8\% improvement in profit, respectively. In contrast, reward shaping, an algorithm that manipulates the reward function, is sensitive to imperfect inputs and leads to a 9.4\% decrease in profit. This highlights the importance of selecting an appropriate mechanism when incorporating imperfect inputs.
\end{abstract}


\end{frontmatter}


\clearpage
\section{Introduction}
\label{sec:introduction}

In response to the threat that climate change poses to global food security, greenhouse production has emerged as a vital strategy for mitigating risks and enhancing food production efficiency worldwide~\cite{Goddek2023, Liao2020, Prakash2021}. Within greenhouse systems, climate control plays a critical role, as it directly influences plant growth by regulating key environmental factors such as temperature, humidity, and CO\textsubscript{2} concentration. However, climate control is also one of the most significant sources of energy consumption in greenhouse operations~\cite{Paris2022}. Looking ahead, greenhouse horticulture faces several pressing challenges, including high energy demands~\cite{WSER}, as well as a shortage of skilled labor and experienced managers~\cite{Christiaensen2020}. Therefore, developing optimal and autonomous climate control systems is essential for improving crop productivity while minimizing energy and resource consumption.

Over the past years, various control methods have been investigated and can be broadly classified into model-based and non-model-based approaches~\cite{vanStraten2010}. Model-based methods rely on a presentation (model) of the greenhouse system (climate, crop, and economic). Among them, model predictive control (MPC) is the most common approach due to its ability to handle the complexity of the greenhouse system~\cite{MAHMOOD2023121190, ZHANG2022}. MPC predicts the future behavior of the system using a model and optimizes its strategy accordingly. In contrast, non-model-based methods rely on knowledge, experience, or data. This type of method includes expert systems~\cite{JACOBSON1989273}, fuzzy logic~\cite{Robles2017}, and genetic algorithms~\cite{pohlheim1996}. However, these methods can’t guarantee optimality as they don’t include dynamic optimization~\cite{vanStraten2010}.

Recently, reinforcement learning (RL) has attracted growing attention in greenhouse climate control. RL is a learning-based method with a model-free nature and adaptive learning capabilities. These features enable RL to perform effectively in complex, error-prone, and uncertain environments~\cite{Lin2020}~\cite{Yizhen2025}, which are characteristic of greenhouse systems. As a data-driven approach, RL can continuously optimize control strategies while adapting to dynamic environmental conditions. Several studies have already evaluated RL in simulated greenhouse settings~\cite{Ajagekar2023, Morcego2023, Zhang2021}, demonstrating its potential for achieving optimal climate control. Despite these promising results, RL applications in greenhouse climate management still face significant challenges. Among the most frequently mentioned limitations are low learning efficiency and limited robustness~\cite{Zhang2021}. Moreover, current RL approaches often exclude growers from the control loop, failing to leverage their domain knowledge and practical experience, which could enhance learning efficiency. This lack of involvement may also lead to growers’ reluctance to adopt RL or other artificial intelligence methods in real-world applications, as they have limited control or understanding of the decision-making process.

Interactive RL is a variant of RL that incorporates human input into the training process. In this approach, the RL agent learns through interactions with both the environment and human feedback. Interactive RL has been tested on classic control tasks like Pac-Man~\cite{Griffith2013}, MountainCar~\cite{Knox2012}, and CartPole~\cite{Knox2012}, and more complicated environments like robotic arm control~\cite{Moreira2020}.
These studies have demonstrated that interactive RL can surpass traditional RL in terms of both training efficiency and overall performance. However, to the best of our knowledge, interactive RL has not yet been explored in the context of greenhouse climate control. In addition, inputs in greenhouse climate control are difficult to perfect and could present a potential challenge for interactive RL.

Given this background, this paper aims to explore the feasibility and performance of applying interactive RL with imperfect inputs to greenhouse climate control. The main contributions are summarized as follows:
\begin{itemize}
\item We developed three representative interactive RL algorithms tailored for greenhouse climate control: policy shaping, which uses grower advice on actions; control sharing, and reward shaping, both of which rely on grower's feedback regarding control performance.
\item We analyzed how input characteristics are often contradicting, and how the trade-offs between them make grower's inputs difficult to perfect. We also discuss the potential shortcomings of different methods inputs can be provided.
\item We proposed a neural network-based approach to enhance the robustness of interactive RL agents under limited input availability by incorporating additional neural networks to aggregate grower's input.
\item We conducted a comprehensive evaluation and comparison of the three interactive RL algorithms in a simulated greenhouse environment, considering the potential variability and characteristics of actual grower input.
\end{itemize}

The remainder of the paper is organized as follows. Section~\ref{sec:model} introduces the simulated greenhouse environment, the baseline RL algorithm, and the background of interactive RL. Section~\ref{sec:input} presents the characteristics of imperfect inputs and how they are simulated. Section~\ref{sec:proposedIRL} introduces the proposed interactive RL algorithms. The evaluation results are then presented and discussed in Section~\ref{sec:result}. Finally, Section~\ref{sec:conclude} draws the conclusion as well as provides future work and recommendations.

\section{Introduction of Greenhouse Environment and RL}
\label{sec:model}
This section begins by introducing the simulated greenhouse environment employed in this paper. It then describes the baseline RL algorithm used, followed by an overview of the interactive RL framework.
\subsection{Greenhouse Environment}
\begin{figure}
    \centering
    \includegraphics[width=1\textwidth]{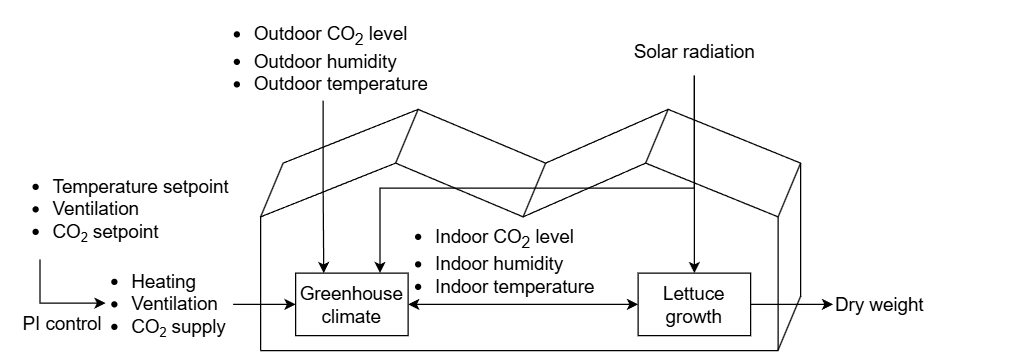}
    \vspace{-10mm}
    \caption{Schematic diagram of the lettuce greenhouse environment~\cite{vanHenten1994}.}
    \label{fig: greenhousemodel}
\end{figure}

This paper investigates a simulated lettuce greenhouse environment under winter weather conditions, as illustrated in Figure~\ref{fig: greenhousemodel}, using the system dynamic model from~\cite{vanHenten1994}. Details of the model are presented in the Appendix. The model is discretized using the Runge-Kutta fourth-order method with a step size of 15 minutes.  Modification to the model input is made as follows: inputs (heating and CO\textsubscript{2} injection rate) are converted into setpoints using proportional-integral (PI) control. These two inputs are converted since providing setpoints as input is more intuitive for growers. However, the ventilation rate is not converted, as it passively influences all climate variables. Table~\ref{tab:environment} summarizes the state and action space of the greenhouse environment. 

The action space of the environment is discrete, with three dimensions: adjusting temperature setpoints, adjusting CO\textsubscript{2} setpoints, and adjusting ventilation rate. This results in a total of 27 possible actions. The control interval of the environment is one hour, while the actuator settings (heating and CO\textsubscript{2} injection rate) are updated every 15 minutes using PI control. The reward function (object function to be maximized) is the profit of the greenhouse minus a penalty, formulated as follows:
\begin{equation}
reward = p_\text{lettuce} \cdot \Delta _\text{dryweight} - ( p_{CO_2} \cdot u_{CO_2} + p_\text{heat} \cdot u_\text{heat})-penalty,
		\label{eq:rewardfunction}
\end{equation}
where \( p_{\text{lettuce}} \) is the price of lettuce (16 Hfl/kg), \( \Delta _\text{dryweight} \) is the change in lettuce dry weight (kg/m\textsuperscript{2}), \( p_{\text{CO}_2} \) is the price of CO\(_2\) supply (0.42 Hfl/kg), \( u_{\text{CO}_2} \) is the usage of CO\(_2\) supply (kg/m\textsuperscript{2}), \( p_{\text{heat}} \) is the price of heating (6.35×10\(^{-9}\) Hfl/J), \( u_{\text{heat}} \) is the usage of heating (J/m\textsuperscript{2}), \(penalty\) is the cost if constraint violated (5.24×10\textsuperscript{-3} Hfl/m\textsuperscript{2}). Note that the price of lettuce, heating, and CO\textsubscript{2} are obtained from~\cite{Morcego2023}. The penalty is applied when indoor climate constraints~\cite{vanHenten1994} (shown in Table~\ref{tab:constraint}) are violated.  The weather data used in the environment and episode length are further detailed in Section~\ref{sec:experimentsettings}.

\begin {table}[t]
\caption{State and action space of the greenhouse environment.}
\vspace{-3mm}
\begin{center}
	\resizebox{.95\textwidth}{!}{\begin{tabular}{l  l  l  l | l  l   }
			\toprule 			
			\multicolumn{4}{c}{\textbf{State space}}&  \multicolumn{2}{c}{\textbf{Action space}}\\
                \midrule
			\textbf{Name}& &   \textbf{Name}& &  \textbf{Dimension} & \textbf{Action} \\
                \midrule
			Outdoor temperature& °C&   Temperature setpoint& °C
&  Temperature setpoint
& -2 °C, 0, +2 °C\\
			Outdoor humidity& \%&   Ventilation rate& mm/s&  Ventilation rate
& 
-0.5 mm/s, 0, +0.5 mm/s\\
                Outdoor CO\textsubscript{2}& ppm&   CO\textsubscript{2} setpoint& ppm&  CO\textsubscript{2} setpoint& $-200~\mathrm{ppm}$, 0, $+200~\mathrm{ppm}$\\
 Indoor temperature& °C
& Hour of the day& & &\\
 
Indoor humidity& \%
& Radiation& W/m\textsuperscript{2}& &\\
 Indoor CO\textsubscript{2}& ppm& Lettuce weight& kg/m\textsuperscript{2}& &\\
 \bottomrule
	\end{tabular}}
	\label{tab:environment}
\end{center}
\end {table} 

\begin {table}[t]
\caption{Constraints for indoor climate.}
\vspace{-3mm}
\begin{center}
	\resizebox{.55\textwidth}{!}{\begin{tabular}{l  l  l  }
			\toprule 			
			\textbf{Indoor climate variable} & \textbf{Lower constraint} & \textbf{Upper constraint} \\
                \midrule
			Temperature (°C)
& 6.5&   40
\\
			CO2 level (ppm)
& 0&   1500
\\
			Relative humidity (\%)& 0&   90
\\
\bottomrule
	\end{tabular}}
	\label{tab:constraint}
\end{center}
\end {table} 
\subsection{Baseline RL Algorithm} 
\label{sec:PPO}
This paper considers the Proximal Policy Optimization (PPO) \cite{Schulman2017} implemented in Stable-Baselines3 \cite{Raffin2021} as the baseline RL algorithm. PPO is selected because its policy is in the form of a probability distribution, which facilitates influencing the action selection of the RL agent. 

\begin{figure}
    \centering
    \includegraphics[width=0.5\textwidth]{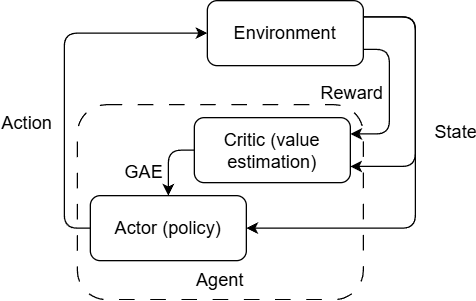}
    \vspace{-3mm}
    \caption{Actor-critic architecture~\cite{Schulman2017}.}
    \label{fig: actorcritic}
\end{figure}

PPO uses the actor-critic architecture shown in Figure~\ref{fig: actorcritic}, where the critic estimates the value function of the state, while the actor represents the policy of the RL agent in the form of a probability distribution. During training, the critic estimates a baseline value for the actor's current policy. Policies that perform better than the current policy are encouraged, while those that perform worse are discouraged. 

Define the temporal difference (TD) at time $t$ as \( \delta_t \) with
\begin{equation}\label{eq:TD}
\delta_t = r_t + \gamma v_\pi(s_{t+1}) - v_\pi(s_t), 		
\end{equation}
where \( r_t \) is the reward at time \( t \), \( v_\pi(s_t) \) is the value of \(s_t\) under policy \( \pi \), \( \gamma\)  is the discount factor.

The level of encouragement or discouragement is based on the generalized advantage estimation (GAE) $A_t$, which estimates how much better or worse a policy is compared to the baseline estimation. $A_t$ is computed by 
\begin{equation}
	\begin{aligned}
        A_t &= \delta_t + (\gamma \lambda) \delta_{t+1} + \cdots + (\gamma \lambda)^{T - t + 1} \delta_{T-1}\\
       & = -v_\pi(s_t) + r_t + (\gamma \lambda) r_{t+1} + \cdots + (\gamma \lambda)^{T - t} v_\pi(s_T),
		\label{eq:advantagefunction}
	\end{aligned}
\end{equation}
where GAE trace parameter \(\lambda\) combines multi-step \( \delta_t \) to balance the trade-off between bias and variance.

\subsection{Interactive RL Backgrounds} 
\label{sec:IRLbackground}
In baseline RL, human involvement is typically limited to the design of the reward function. In contrast, interactive RL allows for a greater degree of human participation throughout the learning process. Interactive RL algorithms have been successfully applied across a variety of control tasks, ranging from simple environments (Pacman, Gridworld)~\cite{Griffith2013, Knox2012} to more complex, realistic scenarios (robot arm control)~\cite{Moreira2020}. These studies have demonstrated that interactive RL can surpass baseline RL in both learning efficiency and overall performance.

In interactive RL, human inputs can be integrated into the learning process through three approaches~\cite{Lin2020, Arzate2020}: reward shaping, policy shaping, and guided exploration. A more detailed overview of these three approaches is presented in Table~\ref{tab:IRLframeworks} and elaborated upon below. 

\begin {table}[t]
\caption{The three approaches of interactive RL.}
\vspace{-3mm}
\begin{center}
	\resizebox{.95\textwidth}{!}{\begin{tabular}{l  l  l>{\raggedright\arraybackslash}p{0.2\linewidth}}
			\toprule 			
\textbf{Approach} & \textbf{Type of Input} & \textbf{Adoption Stage}
&\textbf{Example} \\
                \midrule
			Reward shaping& Feedback&Reward function
&\cite{Knox2010}\\
			Policy shaping& Action advice&Action selection
&\cite{Griffith2013}, \cite{Cederborg2015}
\\
  Guided exploration& Feedback / Action advice& Action selection & \cite{Knox2010}
\\
\bottomrule
	\end{tabular}}
	\label{tab:IRLframeworks}
\end{center}
\end {table} 

\paragraph{Reward Shaping} 
This approach involves modifying the reward function of the RL agent to incorporate additional sources of information. A seminal contribution to this line of work is the TAMER framework proposed by~\cite{Knox2008}, which has been widely adopted in interactive RL. In TAMER, the agent learns a feedback function $F(s,a)$ that estimates human-provided feedback, rather than directly learning the Q-function. Building on this foundation, the authors later introduced the DQN-TAMER framework~\cite{Knox2010}, which integrates both human feedback and environment-generated rewards. In their study, they evaluated eight different methods for combining these two sources of information. Among these, reward shaping emerged as an intuitive and effective strategy, and can be formally expressed as follows:
\begin{equation}
\text{r}'(s, a) = \text{r}(s, a) + \beta \cdot F(s, a),
		\label{eq:rewardshaping}
\end{equation}
where the environment-generated reward $\text{r}(s, a)$ is reshaped by adding a weighted estimated feedback $F(s, a)$.

\paragraph{Policy shaping and guided exploration} 
These two approach both influence the action selection process of the RL agent, but differ in the manner in which this influence is exerted. Policy shaping~\cite{Griffith2013} reshapes the agent’s policy by incorporating:
\begin{equation}
		P(s,a) = 
\begin{cases}
1 - \beta, & a = \pi_{\text{agent}}(s,a) \\
\beta, & a = \dfrac{ \pi_{\text{agent}}(s,a) \cdot \pi_{\text{input}}(s,a) }{ \sum\limits_{a' \in \mathcal{A}} \pi_{\text{agent}}(s,a') \cdot \pi_{\text{input}}(s,a') }
\end{cases},
		\label{eq:policyshaping}
\end{equation}
where $P(s,a)$ is the final probability of selecting action $a$ at state $s$, $\pi_{\text{agent}}(s,a)$ is the probability of 
selecting action $a$ at state $s$ under the agent’s policy, and $\pi_{\text{input}}(s,a)$ is the probability of 
selecting action $a$ at state $s$ under the input's policy. 

Equation \eqref{eq:policyshaping} defines a mixed policy:
with probability $1-\beta$, the agent relies on the agent's policy, and with probability $\beta$, it samples from a mixed distribution that combines both the agent’s and input’s policy. Since this requires that the policy $\pi_{agent}(s)$ be expressed as probabilities of each action, it is typically applied to policy-based algorithms with stochastic policies (PPO, A2C, etc.).

\paragraph{Guided exploration}
In contrast, guided exploration overrides the action selected by the RL agent to guide its exploration~\cite{Arzate2020}, instead of modifying the policy. This makes it applicable to both value-based algorithms and policy-based algorithms.

\section{Grower's Input}\label{sec:input}
Although interactive RL has the potential to outperform baseline RL, it relies on high-quality input to do so~\cite{Bignold2021}. This section begins by outlining the types of input used in interactive RL and the characteristics these inputs must possess. It then examines how, in the context of greenhouse climate control, these desirable characteristics often conflict with one another, creating trade-offs that make it challenging for growers to provide optimal input. Finally, the section details the method employed in this study to simulate imperfect inputs.

\subsection{Human Input in Interactive RL}\label{sec:inputIRL}
In interactive RL, human inputs can be categorized into three types: binary critique feedback, scalar-valued feedback, and action advice~\cite{Arzate2020}. Both binary and scalar-valued feedback relate to the action selected by the agent. Binary feedback simply labels the action as ``good'' or ``bad'', whereas scalar-valued feedback provides a quantitative evaluation of the action. In contrast, action advice specifies either the optimal action or the probability distribution over all possible actions. Beyond the form of information, the scope of input also differs: feedback pertains only to the selected action, while action advice provides information about the entire action space.
In interactive RL, the quality of input is crucial for successive learning~\cite{Bignold2021}. The quality of input can be assessed by seven characteristics~\cite{Bignold2021}, summarized in Table~\ref{tab:inputcharacteristics}.

\begin {table}[t]
\caption{Characteristics of human input~\cite{Bignold2021}.}
\vspace{-3mm}
\begin{center}
	\resizebox{.95\textwidth}{!}{\begin{tabular}{l  l  }
			\toprule 			
			\textbf{Characteristic} & \textbf{Description}\\
                \midrule
			Accuracy
& How appropriate inputs are to the current situation.
\\
			Availability
& The availability of human teachers.
\\
			Concept drift
& The goals or understanding of the environment shift over time.
\\
                Reward bias
& The teaching style of the human teachers, e.g., encouragement and punishment.
\\
 Cognitive bias
& The difference between the RL agent and the human teacher’s goal.
\\
 
Knowledge level
& Knowledge of the environment or information available.
\\
 Latency
& Delay in providing inputs.
\\
\bottomrule
	\end{tabular}}
	\label{tab:inputcharacteristics}
\end{center}
\end {table} 

\subsection{Characteristics of Grower's Input}\label{sec:inputcharacteristics}
In greenhouse climate control, grower's inputs are difficult to perfect due to the trade-offs between these four key characteristics: availability, cognitive bias, latency, and knowledge level. These characteristics are discussed in detail below.

\paragraph{Availability} It refers to the accessibility and willingness of human experts to provide input. While RL agents are often trained in simulated greenhouse environments to reduce time and costs, this setup demands a large volume of input within a short period. Given the repetitive and time-consuming nature of providing inputs, growers may find the task tedious and exhausting, which diminishes their willingness to participate. As a result, querying growers at every step of the training process is impractical, leading to low input availability.
To address this issue, two main strategies are commonly used: applying predefined simple rules or leveraging pre-extracted grower knowledge, such as expert systems. However, both approaches have drawbacks. They may introduce cognitive bias and often lack the depth and nuance of real expert input, thereby reducing the overall knowledge level of the guidance provided.

\paragraph{Cognitive bias} This refers to the misalignment between the RL agent's objective of maximizing final profit and the grower's actual decision-making priorities. Aligning input with the agent’s goal requires complex decision-making under uncertainty, similar to dynamic optimization, which reduces input availability due to the need for frequent grower involvement.
To improve availability, pre-extracted grower knowledge (e.g., expert rules) is often used, but this introduces greater cognitive bias. Growers may prioritize crop physiology over profit due to the long production cycles and economic uncertainties in greenhouse operations. This risk-averse approach aims to ensure high yield and quality, indirectly supporting profitability.
Even when grower input targets profit, it often lacks the adaptability of dynamic optimization, limiting its effectiveness under uncertain and changing conditions~\cite{vanStraten2010}.

\paragraph{Latency}
Input with minimal cognitive bias can also lead to increased latency, which refers to delays in providing input. Latency is particularly problematic for action advice, as it must be delivered during the agent’s action selection process. In greenhouse production, long growing cycles mean that the final profit, often the target of optimization, is only known at the end of the cycle, resulting in inherently high latency. To reduce latency, inputs may instead focus on short-term gains or crop physiology. However, this shift introduces greater cognitive bias.

\paragraph{Knowledge level} It refers to the grower’s understanding of the greenhouse system and the information available for decision-making. Inputs based on high knowledge levels can reduce cognitive bias but often require more time and effort to provide, which decreases availability and increases latency. Conversely, inputs derived from low knowledge levels, such as predefined rules that set allowable ranges for climate variables, offer high availability but tend to exhibit greater cognitive bias.

\subsection{Simulation of Growers' Imperfect Inputs}
\label{sec:inputsimulation}
In this paper, three types of simulated inputs are designed, as detailed in Table~\ref{tab:simulatedinputs}. Both precise action advice and constraint advice are forms of action advice that provide probabilities for each action, but they differ in knowledge level. Feedback with a low knowledge level is excluded, as using feedback to enforce constraints (ranges) is a common practice in RL. 

\begin {table}[t]
\caption{Three types of grower's simulated inputs.}
\begin{center}
	\resizebox{.95\textwidth}{!}{\begin{tabular}{>{\raggedright\arraybackslash}p{0.22\linewidth}>{\raggedright\arraybackslash}p{0.27\linewidth}>{\raggedright\arraybackslash}p{0.27\linewidth}>{\raggedright\arraybackslash}p{0.27\linewidth}}
			\toprule 			
			\textbf{Characteristics}
& \textbf{Feedback} & \textbf{Precise action advice} & \textbf{Constraint advice}\\
                \midrule
			\textit{Input form}
& Binary
(bad action: -1, 
 good action: 1)& Probabilities for action&Probabilities for action\\
			\textit{Cognitive bias}
& Maximize current crop growth
& Maximize current crop growth&Maximize current crop growth
\\
			\textit{Knowledge level}
&   High (consider the optimal values of climate variables)
& High (consider the optimal values of climate variables)&Low (consider the suitable range of climate variables)
\\
\bottomrule
	\end{tabular}}
	\label{tab:simulatedinputs}
\end{center}
\end {table}

All three types of simulated inputs are based on pre-extracted knowledge and inherently contain cognitive bias. They are categorized into two knowledge levels: low and high. Inputs with a low knowledge level specify suitable ranges (constraints) for climate variables, as detailed in Table~\ref{tab:inputrange}. In contrast, inputs with a high knowledge level provide optimal climate variable values aiming at maximizing current crop growth. These optimal values are pre-calculated and approximated using interpolation to facilitate faster input generation.

\begin {table}[t]
\caption{Ranges of climate conditions for constraint advice.}
\vspace{-3mm}
\begin{center}
	\resizebox{.65\textwidth}{!}{\begin{tabular}{l  l  l}
			\toprule 			
			\textbf{Indoor climate variable} & \textbf{Daytime range} & \textbf{Nighttime range}\\
                \midrule
			CO2 setpoint (ppm)& 400 – 1500&300 – 600
\\
			Temperature setpoint (°C)& 10 – 25&6 – 10
\\
\bottomrule
	\end{tabular}}
	\label{tab:inputrange}
\end{center}
\end {table} 

\begin{figure}[t]
    \centering
    \includegraphics[width=1\textwidth]{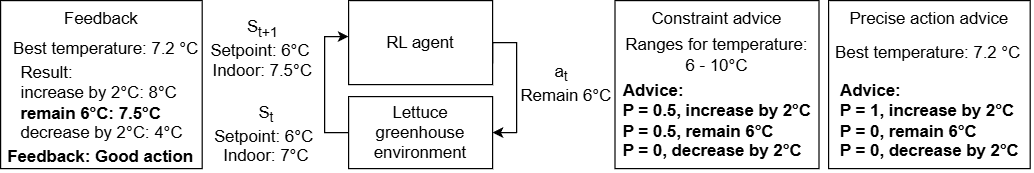}
    \vspace{-10mm}
    \caption{Examples of simulated grower giving input.}
    \label{fig:inputexample}
\end{figure}

Figure~\ref{fig:inputexample} illustrates examples of the three input types, focusing solely on temperature. In constraint advice, actions that keep the temperature within the specified range are assigned equal probabilities of selection. Precise action advice sets the temperature as close as possible to the optimal setpoint. Feedback is provided after the outcome of the action ($S\textsubscript{t+1}$) is observed, evaluating whether the resulting temperature is closer to the optimal value compared to the potential effects of alternative actions.

\section{Interactive RL for Greenhouse Climate Control}\label{sec:proposedIRL}

This section begins by explaining how the three types of inputs described in Section~\ref{sec:inputsimulation} are integrated into PPO. It then introduces a neural network–based approach to enhance the robustness of interactive RL agents when input availability is limited. Finally, it describes the proposed interactive RL algorithm.

\subsection{Incorporating Grower's Inputs} 
In this work, reward shaping and control sharing are employed to incorporate grower feedback, while policy shaping is used to implement the two types of action advices. As introduced in Section~\ref{sec:IRLbackground}, both guided exploration and policy shaping influence action selection. Since PPO is a policy-based algorithm with a stochastic policy, these two approaches are quite similar and are expected to yield comparable performance. Nonetheless, control sharing, a form of guided exploration, is still being evaluated in this work because it allows direct influence on action selection through feedback. Moreover, although using reward shaping to incorporate feedback is intuitive because feedback resembles the human-designed reward function, prior studies suggest that it has limitations since it modifies the reward signal and only affects the policy indirectly~\cite{Knox2010}, whereas control sharing may leverage feedback more effectively.

The used reward shaping and control sharing are adapted from the DQN-TAMER framework~\cite{Knox2008, Knox2010}. Reward shaping modifies the reward according to~\eqref{eq:rewardshaping}, while control sharing influences action selection by overriding the RL agent's policy as below: 
\begin{equation}
	\begin{aligned}
		P(a) = 
\begin{cases}
1 - \beta, & a = \pi_{agent}(s,a) \\
\beta, & a = \arg\max_a \left( F(s,a) \right)
\end{cases}.
		\label{eq:controlsharing}
	\end{aligned}
\end{equation}
Both two approaches employ a weighting factor (likelihood) $\beta$
to regulate the degree of input involvement. Additionally, an auxiliary neural network (the $F$ function) is introduced to estimate feedback in both methods. The 
$F$ function takes state variables as input and outputs estimated feedback values for each possible action.

Policy shaping~\cite{Cederborg2015} is chosen to incorporate both types of action advices. It influences action selection according to~\eqref{eq:policyshaping}. Additionally, a new neural network, 
$\pi_\textsubscript{grower}$, is introduced to estimate the action advice, similar to the $F$ function used in the DQN-TAMER framework.

\subsection{Addressing Limited Availability} 

In all three algorithms, an additional network ($F$ function or $\pi_\textsubscript{grower}$) is used to estimate the input. This network not only facilitates the control sharing method but also enhances the algorithm’s tolerance to limited input availability and high latency. This capability is especially important when inputs must be obtained by querying growers at every step or when providing input demands significant time and effort. 

With the additional network ($F$ function or $\pi_\textsubscript{grower}$), the goal of providing inputs to the agent's training is similar to active learning: maximize the learning efficiency of input estimation with minimal input given. As in active learning, two sampling approaches exist: pool-based and stream-based. This paper only focuses on the pool-based approach for two reasons: First, the pool-based approach has better sample efficiency~\cite{cacciarelli2024}, which suits the needs of limited input availability better. Second, the stream-based approach requires immediate decisions (one chance per step), increasing complexity and potentially the burden of the grower if the final decision is given by the grower.

As in active learning, it is essential to efficiently select which steps to provide input. Within pool-based approach, two selection strategies are compared in this paper: a random strategy and a selective strategy. The random strategy samples input steps randomly, while the selective strategy is inspired by SafeDAgger~\cite{Zhang2016}. In the selective strategy, a new neural network, 
$\pi_\textsubscript{error}$, is introduced to estimate the discrepancy between the simulated input and the model’s estimated input. Inputs are provided only at steps with high predicted discrepancy. Mean squared error is used to quantify the discrepancy for feedback, while Kullback–Leibler (KL) divergence is applied for action advice.

\begin{algorithm}[t]
\caption{The proposed interactive RL algorithm}\label{alg:IRL}
\begin{algorithmic}[1]
\State Initialize PPO agent
\State \textbf{Initialize $F$ function ($\pi_{\text{grower}}$), $\pi_{\text{error}}$ (for the selective strategy), and buffer $D_{\text{input}}$}
\State \textbf{Set hyperparameters: $\beta$, $n$, and $N$}
\While{not reach total timesteps}
    \While{not reach update interval}
        \State \textbf{Interact with the environment using the estimated input}
    \EndWhile
    \State \textbf{Select $n$ steps to provide inputs}
    \State \textbf{Add provided input to $D_{\text{input}}$}
    \State Update the PPO agent
    \State \textbf{Update the $F$ function ($\pi_{\text{grower}}$) and $\pi_{\text{error}}$ (if exists) using $D_{\text{input}}$ for $N$ iterations}
\EndWhile
\end{algorithmic}
\end{algorithm}

\subsection{Interactive RL Implementation} 
The implemented interactive RL algorithms are detailed in Algorithm~\ref{alg:IRL}. Modifications to the baseline PPO framework for incorporating grower inputs and addressing limited input availability are emphasized in bold. In addition to the introduction of auxiliary neural networks, a new buffer, $D_\textsubscript{input}$, is implemented to store data for updating these networks. Furthermore, three new hyperparameters are defined: the weight factor $\beta$, the number of inputs $n$, and the number of update iterations $N$. The weight factor $\beta$ regulates the influence of grower input, as described in \eqref{eq:rewardshaping}, \eqref{eq:policyshaping}, and \eqref{eq:controlsharing}. The parameter $n$ constrains the number of inputs provided during each update interval. The number of update iterations $N$ specifies the number of times the auxiliary networks are updated within each interval.

\section{Results and Discussion}
\label{sec:result}
This section presents a comprehensive evaluation and comparison of the three interactive RL algorithms using simulated imperfect inputs. It begins by describing the experimental setup and training hyperparameters. Subsequently, it evaluates and discusses the performance of the three interactive RL algorithms, the effect of input weight (likelihood), and the influence of input availability.

\subsection{Experiment Settings}\label{sec:experimentsettings}
Hourly weather data from Den Haag~\cite{KNMI} is used to provide outdoor weather for the lettuce greenhouse environment. This dataset includes hourly global radiation, temperature, and humidity. As CO\textsubscript{2} levels are not included, they are assumed to be constant at 400 ppm, since the global average CO\textsubscript{2} is around 400 ppm~\cite{Lan2025}.

For training, each episode spans 14 days (336 steps). The full lettuce growth cycle (56 days) is not used for training due to its long episode, as overly long episodes would cause rollout rewards to reflect outdated training trajectories, and the rollout reward is used as part of the result. However, using an episode length too short might cause challenges for the RL agent to optimize the long-term goal. Through tuning, 14 days were selected for this project.
The initial dry weight is randomly sampled between 3.5 g and 300 g. By doing so, all potential dry weights were included in the shortened episode. Start dates are randomly selected between November and February from the years 2011 to 2020 to reflect typical winter conditions.

For testing, each episode covers a complete 56-day growth cycle, beginning with an initial dry weight of 3.5 g. Weather data from 2021 to 2024 is used. For each year, two test trajectories are selected: one from January 1st to February 25th, and another from December 1st to January 25th of the following year. In total, seven trajectories are used for evaluation.

\begin {table}[t]
\caption{Hyperparameters for all RL agents.}
\vspace{-3mm}
\begin{center}
	\resizebox{.95\textwidth}{!}{\begin{tabular}{>{\raggedright\arraybackslash}p{0.4\linewidth}>{\raggedright\arraybackslash}p{0.15\linewidth}>{\raggedright\arraybackslash}p{0.4\linewidth}>{\raggedright\arraybackslash}p{0.15\linewidth}}
			\toprule 			
			\textbf{Hyperparameters} & \textbf{Values} & \textbf{Hyperparameters} & \textbf{Values}\\
                \midrule
			Learning rate
& 1×10\textsuperscript{-4}&Actor and critic size&512×4
\\
			Step numbers per update
& 2048
&
Gamma (discount factor)&
0.97\\
  Batch size
& 256
& Total training steps
&
500,000\\
\bottomrule
	\end{tabular}}
	\label{tab:hyperparameters1}
\end{center}
\end {table} 

\begin {table}[t]
\caption{Hyperparameters used in interactive RL agents.}
\vspace{-3mm}
\begin{center}
	\resizebox{.95\textwidth}{!}{\begin{tabular}{>{\raggedright\arraybackslash}p{0.45\linewidth}>{\raggedright\arraybackslash}p{0.55\linewidth}}
			\toprule 			
			\textbf{Hyperparameters} & \textbf{Values} \\
                \midrule
			 Sizes of $\pi$\textsubscript{grower} and $\pi$\textsubscript{error} networks &256×3\\
    Number of inputs $n$&2048/1024/512/256/128
\\
   Initial input weight (likelihood) $\beta$&0.5/0.2/0.1/0.05
\\
   Number of iterations $N$&500
\\
   Entropy coefficient
&1×10\textsuperscript{-2}(no inputs, reward shaping, policy shaping(constraint)) / 1×10\textsuperscript{-3} (policy shaping(precise), control sharing)\\
   Learning rate for $\pi$\textsubscript{grower} network & 1×10\textsuperscript{-4}(policy shaping(precise)) / 1×10\textsuperscript{-3} (reward shaping, control sharing)\\
   Learning rate for $\pi$\textsubscript{error} network & 1×10\textsuperscript{-3}\\
   \bottomrule
	\end{tabular}}
	\label{tab:hyperparameters2}
\end{center}
\end {table}

\subsection{RL Agent Training}\label{sec:agenttraining}
The hyperparameters used to train the PPO agent are summarized in Table~\ref{tab:hyperparameters1} and Table~\ref{tab:hyperparameters2}. The value of these hyperparameters is determined by tuning.  The parameters listed in Table~\ref{tab:hyperparameters1} are applied consistently across all RL agents, while those in Table~\ref{tab:hyperparameters2} are specific to the interactive RL algorithms. Among these, the input weight (likelihood) $\beta$ and the number of inputs $n$ are varied to evaluate their impact on the agent’s performance.
In all experiments, the input weight (likelihood) $\beta$ is linearly decreased from its initial value to zero over the first 400,000 steps and remains at zero thereafter. The entropy coefficient, which promotes exploration behavior~\cite{Huang2022}, is also adjusted in interactive RL settings. It is reduced in control sharing and in policy shaping with precise action advice, as both approaches inherently encourage exploration.
To mitigate result variance, each configuration (algorithm, input weight, etc.) is trained using 15 seeds $(100, 200, \dots, 1500)$.

\subsection{ Performance of Interactive RL }\label{sec:irlperformance}
Table~\ref{tab:irlperformance} shows the average cumulative reward of interactive RL algorithms in the test environment. In all experiments, the availability of inputs is full, and the input weights (likelihood) $\beta$ are chosen to be the ones that work best in the test. Additionally, baseline PPO using only simulated feedback as the reward function is also trained and tested,  with its test reward reduced to only 0.71 Hfl/m\textsuperscript{2}. This is expected and indicates that the simulated inputs are far from perfect. The policy shaping using precise action advice performs best in this environment, with 8.4\% improvement. Policy shaping using constraints and control sharing also increase rewards. However, reward shaping reduces the reward, indicating that it is more sensitive to the quality of input. 

\begin {table}[t]
\caption{Average cumulative reward on the test environment.}
\vspace{-3mm}
\begin{center}
	\resizebox{1\textwidth}{!}{\begin{tabular}{>{\raggedright\arraybackslash}p{0.2\linewidth}>{\raggedright\arraybackslash}p{0.18\linewidth}>{\raggedright\arraybackslash}p{0.2\linewidth}>{\raggedright\arraybackslash}p{0.25\linewidth}>{\raggedright\arraybackslash}p{0.25\linewidth}>{\raggedright\arraybackslash}p{0.25\linewidth}>{\raggedright\arraybackslash}p{0.25\linewidth}}
			\toprule 			
			&baseline PPO&PPO with feedback only&Policy shaping (precise) $\beta$=0.2& Policy shaping (constraint) $\beta$=0.1& Control sharing $\beta$=0.05& Reward shaping $\beta$=0.2\\
                \midrule
			Cumulative reward(Hfl/m\textsuperscript{2})& 1.91&0.71&2.07& 1.97& 2.04& 1.73\\
			Relative change to baseline PPO& 0&
-62.8\%&
+8.4\%& +3.1\%& +6.8\%& -9.4\%\\
\bottomrule
	\end{tabular}}
	\label{tab:irlperformance}
\end{center}
\end {table}

\begin{figure}[t]
    \centering
    \includegraphics[width=0.7\textwidth]{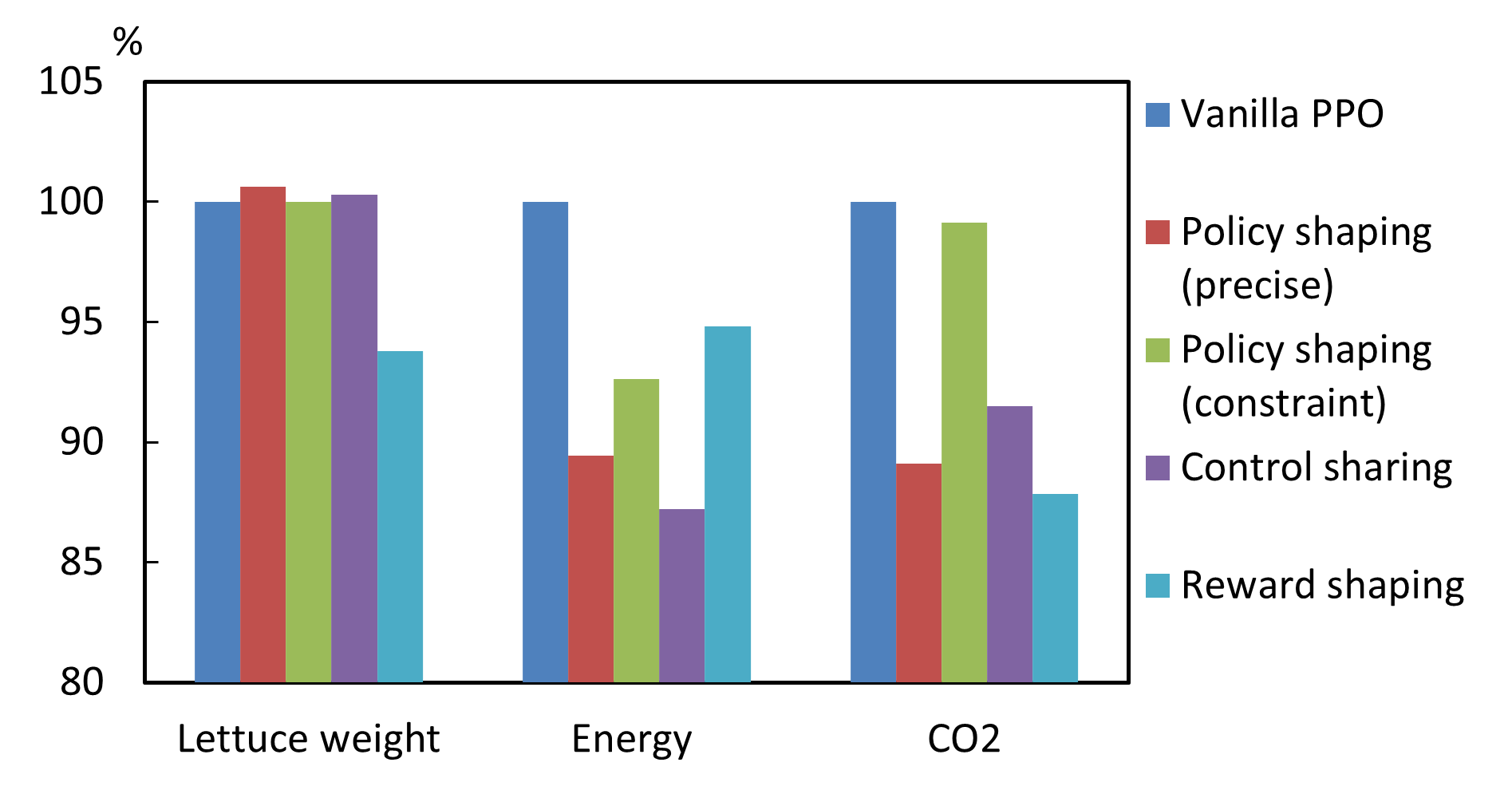}
    \vspace{-6mm}
    \caption{Performance of interactive RL algorithms relative to baseline (Vanilla) PPO.}
    \label{fig: relativeperformance}
\end{figure}

Figure~\ref{fig: relativeperformance} further compares the performance of interactive RL algorithms to baseline PPO. Except for reward shaping, all interactive RL algorithms maintain a similar harvested lettuce weight while reducing energy and CO\textsubscript{2} usage. Policy shaping with precise advice and control sharing each achieves about 10\% reductions in energy and CO\textsubscript{2} usage. Policy shaping with constraint advice only reduces its energy usage by less than 10\%, while its CO\textsubscript{2} usage is only slightly reduced compared with baseline PPO. In contrast, reward shaping lowers both energy and CO\textsubscript{2} usage but also reduces the harvested weight, resulting in a 9.4\% decrease in profit relative to baseline PPO.

\begin{figure}[t]
    \centering
    \includegraphics[width=1\textwidth]{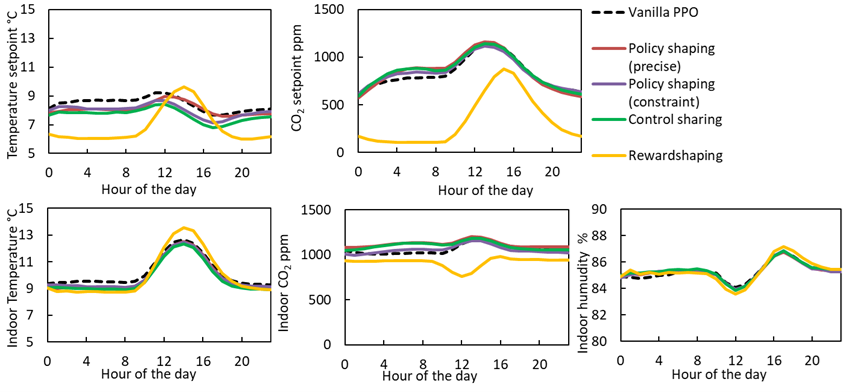}
    \vspace{-12mm}
    \caption{Average setpoints and indoor climate variables over a day.}
    \label{fig: averageclimate}
\end{figure}

Figure~\ref{fig: averageclimate} shows the average setpoints and indoor climate variables for each hour of the day. All interactive RL algorithms, except for reward shaping, show a similar pattern. They maintain lower temperature setpoints and consequently lower indoor temperatures than the baseline PPO. Notably, indoor temperatures and CO\textsubscript{2} setpoints are generally higher than the corresponding indoor values, and high indoor CO\textsubscript{2} and indoor values are observed at night. This is due to several factors: First, the greenhouse environment has no actuator that can actively reduce temperature or CO\textsubscript{2} level. Second, in the later stage of the growing cycle, the plant's respiration can produce a large amount of CO\textsubscript{2}. Third, since there are no violation constraints on high CO\textsubscript{2} level or temperature at night and relative humidity remains within constraints, the agent is not motivated to use ventilation to lower these variables. 

In contrast, reward shaping shows a different pattern. It maintains low temperature setpoints and CO\textsubscript{2} setpoints at night. This is because it receives positive feedback for doing so, without extra cost. During the day, it also maintains low temperature setpoints and CO\textsubscript{2} setpoints, due to the conflict between the reward and the feedback.

The comparison of interactive RL algorithms suggests that even imperfect inputs can improve the RL agent’s performance, except for reward shaping. In contrast, reward shaping follows the feedback's underlying policy and fails to improve the performance of the RL agent. Reward shaping faces the conflict between the reward and the feedback. Manipulating reward functions with imperfect feedback may misguide the learning process~\cite{Knox2012,Knox2010}. Therefore, making it more sensitive to imperfect inputs than other algorithms. 

Policy shaping and control sharing improve performance even when using imperfect input. This is probably because PPO uses GAE to improve its policy. As shown in~\eqref{eq:advantagefunction}, GAE is a type of advantage function that calculates how much better or worse taking an action is compared to $v_\pi(s)$. Both policy shaping and control sharing incorporate inputs by influencing the action selection of the RL agent. Since using feedback as the reward function reduces cumulative rollout rewards, $v_\pi(s)$ can be underestimated when incorporating imperfect inputs, leading to higher GAE estimates, which can accelerate learning. Also, an underestimated $v_\pi(s)$ improves exploration~\cite{Kobayashi2025}, which can help the RL agent explore a better policy.

\subsection{ Impact of Input Weight (Likelihood) }
\label{sec:inputweight}
The input weight (likelihood) significantly influences the performance of interactive RL algorithms. This section focuses on examining its impact in the context of policy shaping with precise action advice, as the effect of likelihood on the other algorithms exhibits similar trends.

\begin{figure}
    \centering
    \includegraphics[width=0.8\textwidth]{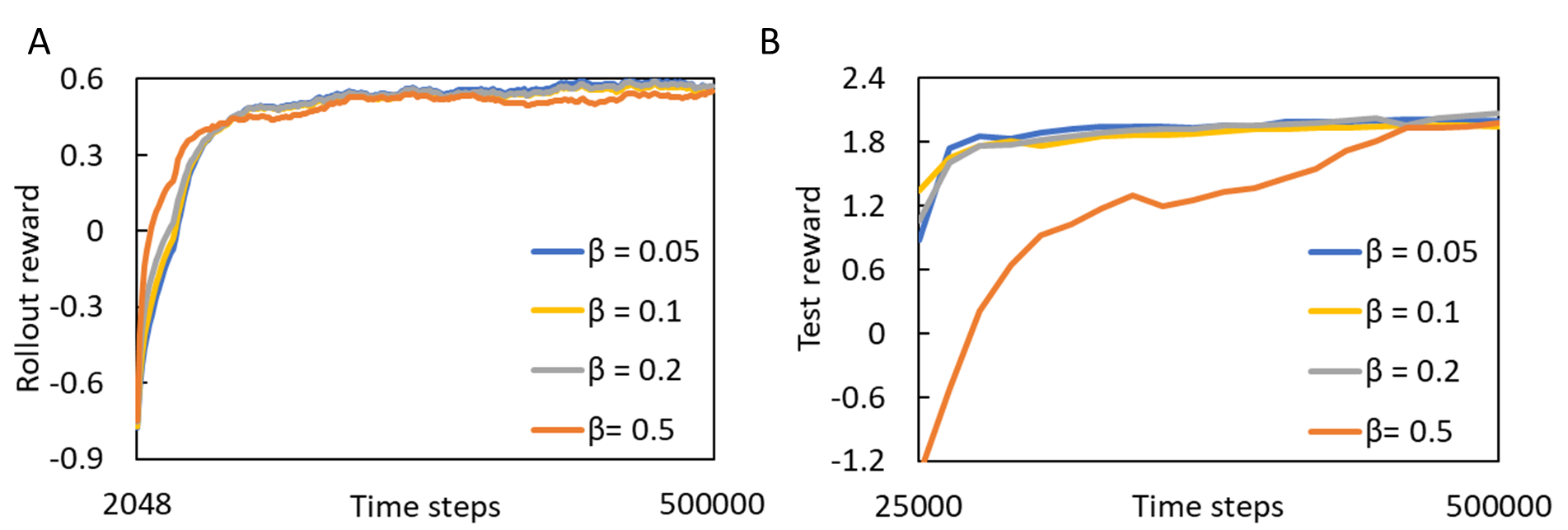}
    \vspace{-5mm}
    \caption{Impact of the likelihood for policy shaping with precise advice. A: rollout reward, B: test reward.}
    \label{fig:changeinreward}
\end{figure}

Figure~\ref{fig:changeinreward} illustrates the evolution of rollout and test rewards during training. The rollout reward refers to the cumulative reward obtained by the RL agent during training and reflects the quality of the agent’s actions within the training environment. As expected, a higher input likelihood accelerates the increase in rollout rewards during the early stages of training. However, this early improvement does not correspond to higher test rewards; instead, it results in lower test performance. Moreover, an excessively high initial likelihood (e.g., $\beta = 0.5$) leads to slightly reduced rollout rewards in the later stages of training, likely due to the way PPO leverages the advantage function to update its policy. In contrast, using a moderately low initial likelihood ($\beta$ in the range of 0.05 to 0.2) yields improvements in both rollout and test rewards.

This result shows that proper input weight (likelihood) is crucial for interactive RL algorithms. Using a proper range of likelihood improves performance. However, a too high likelihood does not improve the performance (accelerates learning), especially in the early stage of training. This is probably also due to PPO using GAE as well.

\begin{figure}
    \centering
    \includegraphics[width=0.8\textwidth]{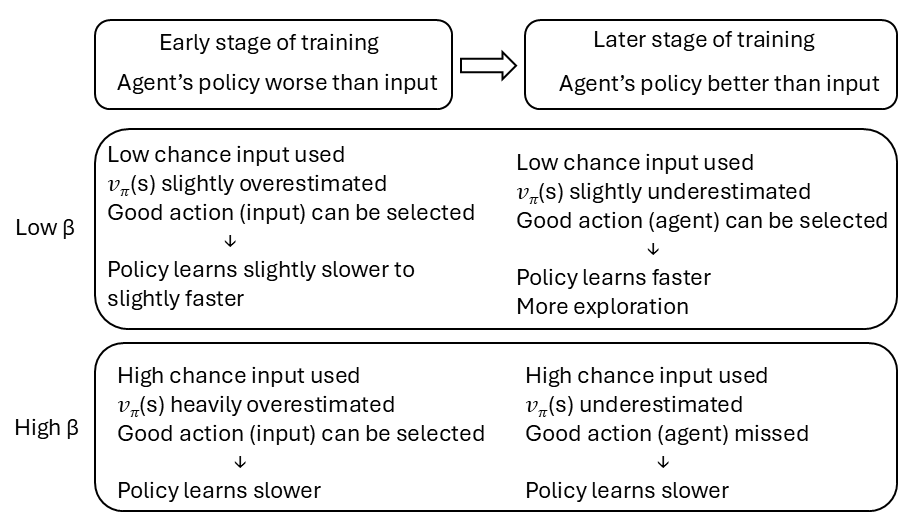}
    \vspace{-5mm}
    \caption{Impact of the likelihood for policy shaping with precise advice. }
    \label{fig: influenceofbeta}
\end{figure}

Figure~\ref{fig: influenceofbeta} shows how the likelihood $\beta$ influences the performance. In the early stage of training, the RL agent struggles to find a policy better than the input. As the likelihood increases, the RL agent takes good actions (input) more frequently, and the critic overestimates $v_\pi(s)$. As shown in Figure~\ref{fig:changeinreward} A, increasing likelihood leads to higher rollout rewards in the early stages of training. This suggests that the RL agent selects actions with high rewards more frequently, and $v_\pi(s)$ can be overestimated. As $v_\pi(s)$ is overestimated, GAE estimates decrease, and the policy learns more slowly.  Therefore, a too high $\beta$ accelerates the increase of rollout rewards but does not translate to high test rewards.  On the other hand, when $\beta$ is low, although $v_\pi(s)$ is slightly overestimated and the policy learns slightly slower, good actions (input) still have a bigger chance of being selected. Therefore, the policy may learn slightly slower to slightly faster (depending on the likelihood $\beta$).

In the later stage of training, the RL agent has already learned a policy better than the input. In this case, incorporating inputs leads to an underestimation of $v_\pi(s)$, which makes the policy learns faster. With a low $\beta$, the agent can still explore and exploit the environment based on its own policy to find a better policy. However, with a high $\beta$, the agent has a smaller chance of selecting actions based on its exploration. Therefore, good actions might be missed, and the policy learns more slowly.

\subsection{Impact of Input Availability}\label{sec:inputavailability}
\begin{figure}
    \centering
    \includegraphics[width=1.0\textwidth]{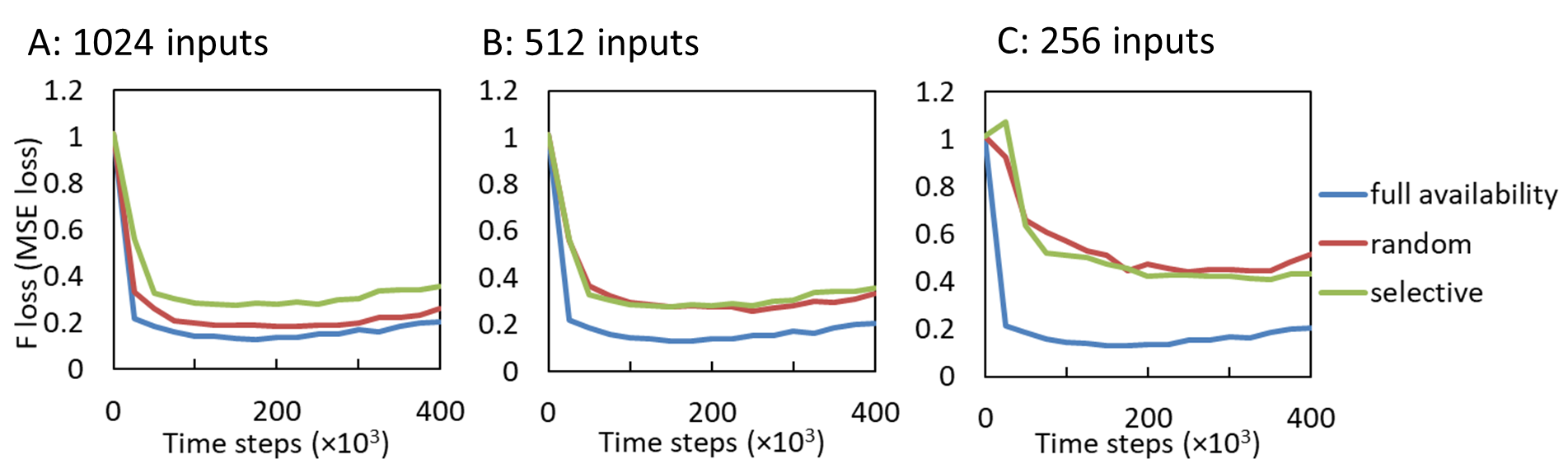}
    \vspace{-10mm}
    \caption{Changes in the average loss of $F$ function (control sharing). }
    \label{fig:floss}
\end{figure}

\begin{figure}
    \centering
    \includegraphics[width=1.0\textwidth]{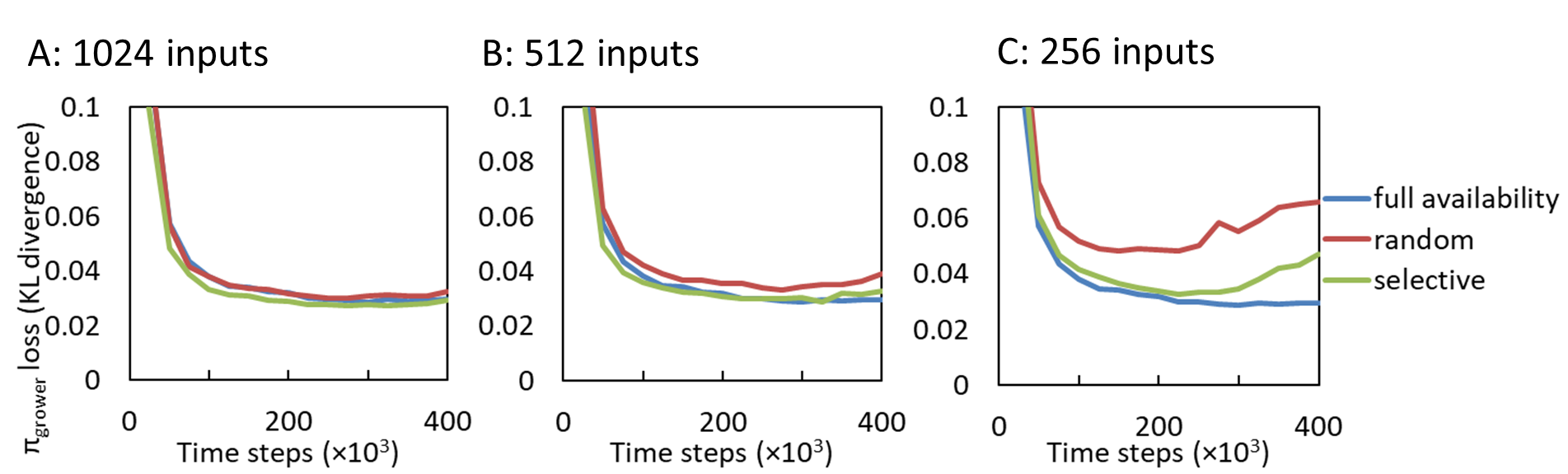}
    \vspace{-10mm}
    \caption{Changes in the average loss of $\pi_\text{grower}$ (policy shaping). }
    \label{fig:piloss}
\end{figure}

Figures~\ref{fig:floss} and \ref{fig:piloss}  show the changes in the average test loss of $F$ function ($\pi_\text{grower}$). Each update interval consists of 2048 steps. When 1024 inputs are provided, this corresponds to input being available for only half of the steps within each interval. The test data is generated by allowing both a well-trained and a poorly trained RL agent to interact with the test environment. Simulated inputs corresponding to these interactions are then collected to form the test dataset. Including both well-trained and poorly trained agents ensures a more balanced and representative test set. In both figures, the time steps are limited to 400,000 instead of 500,000, as estimated inputs are not utilized during the final 100,000 steps, and the $F$ function ($\pi_\text{grower}$) is no longer updated during this period.

Figure~\ref{fig:floss} depicts the changes in the loss of the $F$ function (control sharing). A reduction in the number of inputs leads to a significant increase in the loss of the $F$ function. Additionally, the selective strategy does not reduce the loss compared to the random strategy, except when 256 inputs are used per update interval. Figure~\ref{fig:piloss} presents the changes in the loss of $\pi_\text{grower}$ (policy shaping). At the initial time step, the loss is 2.23 and then decreases rapidly. Across all three cases, the selective strategy either does not increase the loss (for 1024 and 512 inputs) or only slightly increases it (for 256 inputs) during the early stages of training. Moreover, the selective strategy consistently results in a lower loss of $\pi_\text{grower}$ compared to the random strategy across all input levels. It is also noteworthy that the loss increases during the later stages of training in the cases with 512 and 256 inputs per update interval, regardless of the strategy employed.

These results indicate that the impact of limited input availability differs between the two types of inputs. For feedback under control sharing, neither the selective nor the random strategy is effective in maintaining a low loss for the $F$ function. In contrast, for action advice, the selective strategy consistently outperforms the random strategy in reducing the loss of $\pi_\text{grower}$. This distinction suggests that action advice is less sensitive to limited input availability than feedback. The effectiveness of the selective strategy in the context of action advice further implies that selecting inputs based on estimated discrepancy yields more informative updates, thereby enhancing the learning efficiency of $\pi_\text{grower}$. Additionally, the increase in loss observed during the later stages of training in Figure~\ref{fig:piloss} C suggests that $\pi_\text{grower}$ may be overfitting, which could hinder its generalization to unseen data. Future work may address this issue by incorporating regularization techniques or applying early stopping to improve generalization performance.

\section{Conclusion}
\label{sec:conclude}
Interactive RL combines human (grower) input with the RL agent’s learning to improve the overall performance of the RL agent. However, as it has not yet been applied to greenhouse climate control and faces challenges of imperfect inputs. This work investigated the possibility and performance of applying interactive RL with imperfect inputs in greenhouse climate control. The key contributions and findings are as follows: 1) Three interactive RL algorithms, policy shaping, control sharing, and reward shaping, are proposed and evaluated in a simulated greenhouse environment with simulated imperfect inputs. Simulation results indicate that interactive RL, even with imperfect grower inputs, can enhance agent performance, provided the interactive approach is appropriately selected and designed. As a consequence of using GAE in training, methods influencing action selection (policy shaping and control sharing) are more robust to the incorporation of imperfect input, provided the level of incorporation is properly tuned. 2) This work also analyzed the trade-off between key characteristics of imperfect input, including availability, cognitive bias, latency, and knowledge level. The drawbacks of strategies such as pre-extracting grower knowledge and direct grower interaction were also discussed. 3) Moreover, a neural network–based method to address limited input availability is proposed. This method uses a neural network to estimate inputs and convert providing inputs to a process similar to pool-based active learning. The test results indicate that action advice is more robust to limited availability than feedback, particularly when using the discrepancy-based selection strategy.

Despite the contributions, this work has certain limitations that can be addressed in future research. Only PPO is selected as the baseline RL algorithm. As discussed, GAE used in PPO might be the reason why policy shaping and control sharing are robust to imperfect inputs and sensitive to the level of input incorporation. However, this may not be true for other RL algorithms. The performance of RL algorithms without using the advantage function may not be improved, which is not explored in this work. Additionally, in the DQN-TAMER framework~\cite{Knox2012}, a too-high level of input incorporation also led to worse performance, but a higher level of input incorporation is tolerated.  

Future work should explore interactive RL in continuous action spaces and its integration with other RL algorithms. Additionally, it is important to understand grower's control strategy and decision-making process. For example, inverse RL can be used to understand grower's underlying intention and goal. This not only helps with providing better inputs, but can also improve reward design for RL agents. It also enables growers to gain a clearer understanding of their intentions and strategies, helping them to prioritize decisions more effectively.


\section*{APPENDIX}
\noindent The lettuce greenhouse model is described by the following equations:
\begin{equation*}
	\begin{aligned}
		\frac{\text{d} x_1(t)}{\text{d}t} &=  p_{1,1} \phi_{\text{phot,c}}(t) -  p_{1,2} x_1(t) 2^{x_3(t)/10-5/2},  \\
		\frac{\text{d} x_2(t)}{\text{d}t} &= \frac{1}{p_{2,1}} \left( -\phi_{\text{phot,c}}(t) + p_{2,2} x_1(t) 2^{x_3(t)/10-5/2} + u_1(t) 10^{-6} - \phi_{\text{vent,c}}(t) \right), \\
		\frac{\text{d} x_3(t)}{\text{d}t} &= \frac{1}{p_{3,1}} \left( u_3(t) - (p_{3,2} u_2(t)10^{-3} +p_{3,3}) (x_3(t)-d_3(t)) + p_{3,4} d_1(t) \right), \\
		\frac{\text{d} x_4(t)}{\text{d}t} &= \frac{1}{p_{4,1}} \big( \phi_{\text{transp,h}}(t) - \phi_{\text{vent,h}}(t) \big), \\
	\end{aligned}
\end{equation*}
with
\begin{equation*}
	\begin{aligned}
		\phi_{\text{phot,c}}(t) ={}& (1-\text{exp} (-p_{1,3}x_1(t) ) ) [p_{1,4}d_1(t) (-p_{1,5}x_3(t)^2+ p_{1,6} x_3(t) - p_{1,7}) \\ 
		& \times ( x_2(t)- p_{1,8})] /\varphi(t),\\
		\varphi(t)              ={}& p_{1,4}d_1(t)+\big(-p_{1,5}x_3(t)^2+p_{1,6} x_3(t) - p_{1,7} \big) \big( x_2(t)- p_{1,8}\big), \\
		\phi_{\text{vent,c}}(t) ={}& \big(u_2(t)10^{-3}+p_{2,3}\big)\big(x_2(t)-d_2(t)\big),\\
		\phi_{\text{vent,h}}(t) ={}& \big(u_2(t)10^{-3}+p_{2,3}\big)\big(x_4(t)-d_4(t)\big), \\
		\phi_{\text{transp,h}}(t) ={}& p_{4,2} (1-\text{exp} (-p_{1,3}x_1(t) ) ) \\ 
		& \times \left( \frac{p_{4,3}}{p_{4,4}(x_3(t)+p_{4,5})}  \text{exp} \left( \frac{p_{4,6} x_3(t)}{x_3(t) + p_{4,7}} \right) - x_4(t) \right),	\\
	\end{aligned}
\end{equation*}  
\begin{equation*}
	\begin{aligned}        
        u_1(t) ={}& p_{5,1} (u_{CO_2}(t)- y_2(t)) + p_{5,2} \int_{t - 1\,\text{hour}}^{t} (u_{CO_2}(\tau)- y_2(\tau)) \, d\tau,\\
        u_3(t) ={}& p_{5,3} (u_{temp}(t)- y_3(t)) + p_{5,4} \int_{t - 1\,\text{hour}}^{t} (u_{temp}(\tau)- y_3(\tau)) \, d\tau,\\
	\end{aligned}
\end{equation*}
where the meaning of variables is shown in Table~\ref{tab:signalsModel}. Furthermore, $\phi_{\text{phot,c}}(t)$, $\phi_{\text{vent,c}}(t),$ $\phi_{\text{transp,h}}(t)$ and $\phi_{\text{vent,h}}(t)$ are the gross canopy photosynthesis rate, mass exchange of CO$_2$ through the vents, canopy transpiration and mass exchange of H$_2$O through the vents, respectively. The measurement equation is defined as
\begin{equation*}
	\begin{aligned}
		y_1(t) &= 10^3 \cdot x_1(t)  	&\quad \text{g} \cdot \text{m}^{-2}, \\
		y_2(t) &= \frac{10^6 \cdot p_{2,4} ( x_3(t) + p_{2,5} ) } {p_{2,6} p_{2,7}} \cdot x_2(t),  	&\quad \text{ppm} , \\
		y_3(t) &= x_3(t), 	&\quad \text{°C}, \\
		y_4(t) &=  \frac{10^2 \cdot p_{2,4} \big( x_3(t)+ p_{2,5} \big)}{11 \cdot \text{exp}\Big( \frac{p_{4,8} x_3(t)}{x_3(t)+p_{4,9}} \Big)} \cdot x_4(t), 	&\quad \text{\%},
	\end{aligned}
\end{equation*}
The model parameters $p_{i,j}$ are defined and presented in Table~\ref{tab:model_parameters}. 

\begin {table}[th]
\caption{Meaning of the lettuce greenhouse model parameters.}
\vspace{-8mm}
\begin{center}
	\resizebox{1\textwidth}{!}{\begin{tabular}{l  l  |l  l   }
			\toprule 			
			$x_1(t)$ & dry-weight (kg/m$^2$) &   $d_1(t)$ & radiation (W/m$^2$) 
\\
			$x_2(t)$ & indoor CO$_2$ (kg/m$^3$) &   $d_2(t)$ & outdoor CO$_2$ (kg/m$^3$) 
\\
			$x_3(t)$ & indoor temperature ($^\circ$C)&   $d_3(t)$ & outdoor temperature ($^\circ$C)\\
			$x_4(t)$ & indoor humidity (\%) &   $d_4(t)$ & outdoor humidity (kg/m$^3$)\\
            \midrule
             $u_1(t)$ &  CO$_2$ injection (mg/m$^2$/s) &  $y_1(t)$&  dry-weight (kg/m$^2$)\\
             $u_2(t)$ &  ventilation (mm/s) &   $y_{2}(t)$ &  indoor CO$_2$ (ppm)
\\
             $u_3(t)$&  heating (W/m$^2$) &   $y_{3}(t)$ & indoor temperature ($^\circ$C)\\
 $u_{CO_2}(t)$& CO$_2$ setpoint (ppm)
& $y_{4}(t)$ &indoor humidity (\%) \\
 $u_{temp}(t)$& temperature setpoint ($^\circ$C)& &\\
 \bottomrule
	\end{tabular}}
	\label{tab:signalsModel}
\end{center}
\end {table} 

\begin{table}[th]
	\center
	\caption{\label{tab:model_parameters}Model parameter values~\cite{vanHenten1994}.}
    \vspace{-4mm}
	\resizebox{1\textwidth}{!}{	\begin{tabular}{ cc|cc|cc|cc|cc} 
    \toprule 
			parameter & value & parameter & value & parameter & value & parameter & value & parameter&value\\
			\hline
			$p_{1,1}$ & 0.544 & $p_{2,1}$ 	& 4.1 			& $p_{3,1}$ & 3$\cdot 10^{4}$ & $p_{4,1}$ 	& 4.1  & $p_{5,1}$&0.05\\ 
			$p_{1,2}$ & 2.65 $\cdot 10^{-7}$ & $p_{2,2}$ 	& 4.87 $\cdot 10^{-7}$ 			& $p_{3,2}$ & 1290 & $p_{4,2}$ & 0.0036  & $p_{5,2}$&3 $\cdot 10^{-6}$\\ 
			$p_{1,3}$ & 53 & $p_{2,3}$ 	& 7.5 $\cdot 10^{-6}$ 			& $p_{3,3}$ & 6.1 & $p_{4,3}$ 	& 9348  & $p_{5,3}$&55\\ 
			$p_{1,4}$ & 3.55 $\cdot 10^{-9}$ &	$p_{2,4}$	& 8.31  		& $p_{3,4}$ & 0.2 & $p_{4,4}$ 	& 8314  & $p_{5,4}$&2.5 $\cdot 10^{-2}$\\ 
			$p_{1,5}$ & 5.11 $\cdot 10^{-6}$ & $p_{2,5}$	& 273.15		&  			&   & $p_{4,5}$ 	& 273.15  & &\\ 
			$p_{1,6}$ & 2.3 $\cdot 10^{-4}$ & $p_{2,6}$	& 101325		&  			&   & $p_{4,6}$ 	& 17.4  & &\\ 
			$p_{1,7}$ & 6.29 $\cdot 10^{-4}$ & $p_{2,7}$	& 0.044			&  			&   & $p_{4,7}$ 	& 239  & &\\ 
			$p_{1,8}$ & 5.2 $\cdot 10^{-5}$ & 			&   			&  			&   &  $p_{4,8}$  	&   17.269   & &\\ 
			&   & 			&   			&  			&   							&  $p_{4,9}$	&  238.3   & &\\
            \bottomrule
	\end{tabular}}
\end{table}	

\bibliography{main}

\begin{thebibliography}{10}
\expandafter\ifx\csname url\endcsname\relax
  \def\url#1{\texttt{#1}}\fi
\expandafter\ifx\csname urlprefix\endcsname\relax\def\urlprefix{URL }\fi
\expandafter\ifx\csname href\endcsname\relax
  \def\href#1#2{#2} \def\path#1{#1}\fi

\bibitem{Goddek2023}
S.~Goddek, O.~Körner, K.~J. Keesman, M.~A. Tester, R.~Lefers, L.~Fleskens, A.~Joyce, E.~van Os, A.~Gross, R.~Leemans, How greenhouse horticulture in arid regions can contribute to climate-resilient and sustainable food security, Global Food Security 38 (2023) 100701.

\bibitem{Liao2020}
P.-A. Liao, J.-Y. Liu, L.-C. Sun, H.-H. Chang, Can the adoption of protected cultivation facilities affect farm sustainability?, Sustainability 12~(23) (2020) 9970.

\bibitem{Prakash2021}
P.~Prakash, P.~Kumar, A.~Kar, P.~Kishore, A.~K. Singh, S.~Immanuel, Protected cultivation in maharashtra: determinants of adoption, constraints, and impact, Agricultural Economics Research Review 34~(2) (2021).

\bibitem{Paris2022}
B.~Paris, F.~Vandorou, A.~T. Balafoutis, K.~Vaiopoulos, G.~Kyriakarakos, D.~Manolakos, G.~Papadakis, Energy use in greenhouses in the eu: A review recommending energy efficiency measures and renewable energy sources adoption, Applied Sciences 12~(10) (2022) 5150.

\bibitem{WSER}
W.~Social, E.~Research, \href{https://agrimatie.nl/SectorResultaat.aspx?subpubID=2232&sectorID=2240}{Agro \& food portal}.
\newline\urlprefix\url{https://agrimatie.nl/SectorResultaat.aspx?subpubID=2232&sectorID=2240}

\bibitem{Christiaensen2020}
L.~Christiaensen, Z.~Rutledge, J.~E. Taylor, The future of work in agriculture: Some reflections, World Bank Policy Research Working Paper~(9193) (2020).

\bibitem{vanStraten2010}
G.~Van~Straten, G.~van Willigenburg, E.~van Henten, R.~van Ooteghem, Optimal control of greenhouse cultivation, CRC press, 2010.

\bibitem{MAHMOOD2023121190}
F.~Mahmood, R.~Govindan, A.~Bermak, D.~Yang, T.~Al-Ansari, \href{https://www.sciencedirect.com/science/article/pii/S0306261923005548}{Data-driven robust model predictive control for greenhouse temperature control and energy utilisation assessment}, Applied Energy 343 (2023) 121190.
\newblock \href {https://doi.org/https://doi.org/10.1016/j.apenergy.2023.121190} {\path{doi:https://doi.org/10.1016/j.apenergy.2023.121190}}.
\newline\urlprefix\url{https://www.sciencedirect.com/science/article/pii/S0306261923005548}

\bibitem{ZHANG2022}
M.~Zhang, T.~Yan, W.~Wang, X.~Jia, J.~Wang, J.~J. Klemeš, \href{https://www.sciencedirect.com/science/article/pii/S1364032122004981}{Energy-saving design and control strategy towards modern sustainable greenhouse: A review}, Renewable and Sustainable Energy Reviews 164 (2022) 112602.
\newblock \href {https://doi.org/https://doi.org/10.1016/j.rser.2022.112602} {\path{doi:https://doi.org/10.1016/j.rser.2022.112602}}.
\newline\urlprefix\url{https://www.sciencedirect.com/science/article/pii/S1364032122004981}

\bibitem{JACOBSON1989273}
B.~Jacobson, P.~H. Jones, J.~Jones, J.~Paramore, \href{https://www.sciencedirect.com/science/article/pii/0168169989900185}{Real-time greenhouse monitoring and control with an expert system}, Computers and Electronics in Agriculture 3~(4) (1989) 273--285.
\newblock \href {https://doi.org/https://doi.org/10.1016/0168-1699(89)90018-5} {\path{doi:https://doi.org/10.1016/0168-1699(89)90018-5}}.
\newline\urlprefix\url{https://www.sciencedirect.com/science/article/pii/0168169989900185}

\bibitem{Robles2017}
C.~Robles~Algarín, J.~Callejas~Cabarcas, A.~Polo~Llanos, \href{https://www.mdpi.com/2079-9292/6/4/71}{Low-cost fuzzy logic control for greenhouse environments with web monitoring}, Electronics 6~(4) (2017).
\newblock \href {https://doi.org/10.3390/electronics6040071} {\path{doi:10.3390/electronics6040071}}.
\newline\urlprefix\url{https://www.mdpi.com/2079-9292/6/4/71}

\bibitem{pohlheim1996}
H.~Pohlheim, A.~Hei{\ss}ner, Optimal control of greenhouse climate using genetic algorithms, in: Proceedings of the MENDEL, Vol.~96, 1996, pp. 112--119.

\bibitem{Lin2020}
J.~Lin, Z.~Ma, R.~Gomez, K.~Nakamura, B.~He, G.~Li, A review on interactive reinforcement learning from human social feedback, IEEE Access 8 (2020) 120757--120765.
\newblock \href {https://doi.org/10.1109/ACCESS.2020.3006254} {\path{doi:10.1109/ACCESS.2020.3006254}}.

\bibitem{Yizhen2025}
Y.~Meng, C.~Liu, J.~Zhao, J.~Huang, G.~Jing, Stackelberg game-based anti-disturbance control for unmanned surface vessels via integrative reinforcement learning, Intelligence and Robotics 5~(1) (2025) 88 -- 104.

\bibitem{Ajagekar2023}
A.~Ajagekar, N.~S. Mattson, F.~You, Energy-efficient ai-based control of semi-closed greenhouses leveraging robust optimization in deep reinforcement learning, Advances in Applied Energy 9 (2023) 100119.

\bibitem{Morcego2023}
B.~Morcego, W.~Yin, S.~Boersma, E.~Van~Henten, V.~Puig, C.~Sun, Reinforcement learning versus model predictive control on greenhouse climate control, Computers and Electronics in Agriculture 215 (2023) 108372.

\bibitem{Zhang2021}
W.~Zhang, X.~Cao, Y.~Yao, Z.~An, X.~Xiao, D.~Luo, Robust model-based reinforcement learning for autonomous greenhouse control, in: Asian Conference on Machine Learning, PMLR, 2021, pp. 1208--1223.

\bibitem{Griffith2013}
S.~Griffith, K.~Subramanian, J.~Scholz, C.~L. Isbell, A.~L. Thomaz, Policy shaping: Integrating human feedback with reinforcement learning, Advances in neural information processing systems 26 (2013).

\bibitem{Knox2012}
W.~B. Knox, P.~Stone, Reinforcement learning from simultaneous human and mdp reward, in: AAMAS, Vol. 1004, Valencia, 2012, pp. 475--482.

\bibitem{Moreira2020}
I.~Moreira, J.~Rivas, F.~Cruz, R.~Dazeley, A.~Ayala, B.~Fernandes, Deep reinforcement learning with interactive feedback in a human–robot environment, Applied Sciences 10~(16) (2020) 5574.

\bibitem{vanHenten1994}
E.~Van~Henten, Greenhouse climate management: an optimal control approach, Wageningen University and Research, 1994.

\bibitem{Schulman2017}
J.~Schulman, F.~Wolski, P.~Dhariwal, A.~Radford, O.~Klimov, Proximal policy optimization algorithms, arXiv preprint arXiv:1707.06347 (2017).

\bibitem{Raffin2021}
A.~Raffin, A.~Hill, A.~Gleave, A.~Kanervisto, M.~Ernestus, N.~Dormann, Stable-baselines3: Reliable reinforcement learning implementations, Journal of Machine Learning Research 22~(268) (2021) 1--8.

\bibitem{Arzate2020}
C.~Arzate~Cruz, T.~Igarashi, A survey on interactive reinforcement learning: Design principles and open challenges, in: Proceedings of the 2020 ACM designing interactive systems conference, 2020, pp. 1195--1209.

\bibitem{Knox2010}
W.~B. Knox, P.~Stone, Combining manual feedback with subsequent mdp reward signals for reinforcement learning, in: AAMAS, Vol.~10, 2010, pp. 5--12.

\bibitem{Cederborg2015}
T.~Cederborg, I.~Grover, C.~L. Isbell~Jr, A.~L. Thomaz, Policy shaping with human teachers, in: IJCAI, 2015, pp. 3366--3372.

\bibitem{Knox2008}
W.~B. Knox, P.~Stone, Tamer: Training an agent manually via evaluative reinforcement, in: 2008 7th IEEE international conference on development and learning, IEEE, 2008, pp. 292--297.

\bibitem{Bignold2021}
A.~Bignold, F.~Cruz, R.~Dazeley, P.~Vamplew, C.~Foale, An evaluation methodology for interactive reinforcement learning with simulated users, Biomimetics 6~(1) (2021) 13.

\bibitem{cacciarelli2024}
D.~Cacciarelli, M.~Kulahci, Active learning for data streams: a survey, Machine Learning 113~(1) (2024) 185--239.

\bibitem{Zhang2016}
J.~Zhang, K.~Cho, Query-efficient imitation learning for end-to-end autonomous driving, arXiv preprint arXiv:1605.06450 (2016).

\bibitem{KNMI}
KNMI, \href{https://www.knmi.nl/nederland-nu/klimatologie/uurgegevens}{Knmi - hourly weather data in the netherlands}.
\newline\urlprefix\url{https://www.knmi.nl/nederland-nu/klimatologie/uurgegevens}

\bibitem{Lan2025}
X.~Lan, P.~Tans, K.~Thoning, Trends in globally-averaged co2 determined from noaa global monitoring laboratory measurements.
\newblock \href {https://doi.org/https://doi.org/10.15138/9N0H-ZH07} {\path{doi:https://doi.org/10.15138/9N0H-ZH07}}.

\bibitem{Huang2022}
S.~Huang, R.~F.~J. Dossa, A.~Raffin, A.~Kanervisto, W.~Wang, \href{https://iclr-blog-track.github.io/2022/03/25/ppo-implementation-details/}{The 37 implementation details of proximal policy optimization}, in: ICLR Blog Track, 2022, https://iclr-blog-track.github.io/2022/03/25/ppo-implementation-details/.
\newline\urlprefix\url{https://iclr-blog-track.github.io/2022/03/25/ppo-implementation-details/}

\bibitem{Kobayashi2025}
T.~Kobayashi, Intentionally-underestimated value function at terminal state for temporal-difference learning with mis-designed reward, Results in Control and Optimization 18 (2025) 100530.

\end{thebibliography}

\end{document}